\documentclass[a4paper,conference]{IEEEtran}
\usepackage[pdftex]{graphicx}
\usepackage{amsmath}
\usepackage{hyperref}
\usepackage{eso-pic}
\usepackage{fancyhdr}
\fancypagestyle{firstpage}{
\lhead{25th International Conference on Pattern Recognition (ICPR), Milan Italy, January 10-15 2021}
\rhead{}
}

\begin{document}

\title{Pixel-based Facial Expression Synthesis}
\author{\IEEEauthorblockN{Arbish Akram and Nazar Khan}
\IEEEauthorblockA{
Punjab University College of Information Technology\\
Lahore, Pakistan \\
Email: \{arbishakram, nazarkhan\}@pucit.edu.pk}
}

\maketitle

\begin{abstract}
Facial expression synthesis has achieved remarkable advances with the advent of Generative Adversarial Networks (GANs). However, GAN-based approaches mostly generate photo-realistic results as long as the testing data distribution is close to the training data distribution. The quality of GAN results significantly degrades when testing images are from a slightly different distribution. Moreover, recent work has shown that facial expressions can be synthesized by changing localized face regions. In this work, we propose a pixel-based facial expression synthesis method in which each output pixel observes only one input pixel. The proposed method achieves good generalization capability by leveraging only a few hundred training images. Experimental results demonstrate that the proposed method performs comparably well against state-of-the-art GANs on in-dataset images and significantly better on out-of-dataset images. In addition, the proposed model is two orders of magnitude smaller which makes it suitable for deployment on resource-constrained devices. 
\end{abstract}
\IEEEpeerreviewmaketitle

\thispagestyle{firstpage}

\section{Introduction}
Facial expression synthesis (FES) has received considerable attention in recent years. FES aims to generate  identity-preserving faces conditioned on a specific facial expression. It is a useful task for the animation of characters and avatars in video games and animated movies. It can be used to control interactive virtual assistants or to help in video surveillance across expressions. 
While humans can easily visualize/synthesize a person's face with a different facial expression, this task is captivating and challenging for computers. 

Early methods for FES extract facial geometric features \cite{zhang2005geometry} or 3D meshes \cite{queiroz2017framework} from images to transfer facial expressions on new faces or animated avatars. Traditional statistical learning-based methods including Active Appearance models \cite{cootes2001active}, Active Shape models \cite{cootes1992active}, 3D warping-based methods \cite{blanz1999morphable}, constrained local methods \cite{cristinacce2006feature} and bilinear kernel-based methods \cite{huang2010bilinear} are also used for synthesizing realistic facial expressions. Since this synthesis task learns the mapping between high-dimensional input and output spaces, models such as regression or neural networks require a lot of data to learn the optimal parameters to approximate this mapping. The number of parameters of these models grows exponentially as the image size increases. Therefore, image synthesis in general and FES in particular, has received less attention as compared to recognition based tasks.

Recently, significant progress has been achieved in image synthesis related areas with the emergence of Generative Adversarial Networks (GANs) \cite{goodfellow-2014, mirza2014conditional}. GANs have proven effective for many image synthesis related tasks including image generation \cite{goodfellow-2014,radford-2015,perarnau-2016,karras2019style, karras2020analyzing, karras2017progressive}, image-to-image translation \cite{zhu2016generative, isola-2016}, image super-resolution \cite{ledig2017photo,hui2018fast}, text-based image synthesis \cite{zhang-2017stackgan} and many others \cite{jin2017towards, wu2016learning, gansynth}. GANs have also obtained remarkable progress in face-related tasks, such as facial attributes editing \cite{perarnau-2016, shen-2016}, face aging \cite{zhang-2017} and facial expression synthesis \cite{choi-2017, pumarola2018ganimation}. These GANs generate better results on in-dataset images as compared to out-of-dataset images. By in-dataset, we mean that training and testing images are taken from the same dataset that follows a certain distribution. In contrast, out-of-dataset images are unseen images that follow a significantly different distribution from the training distribution.

\begin{figure*}[t]
    \centering
    \includegraphics[width=1.01\textwidth,height=8cm]{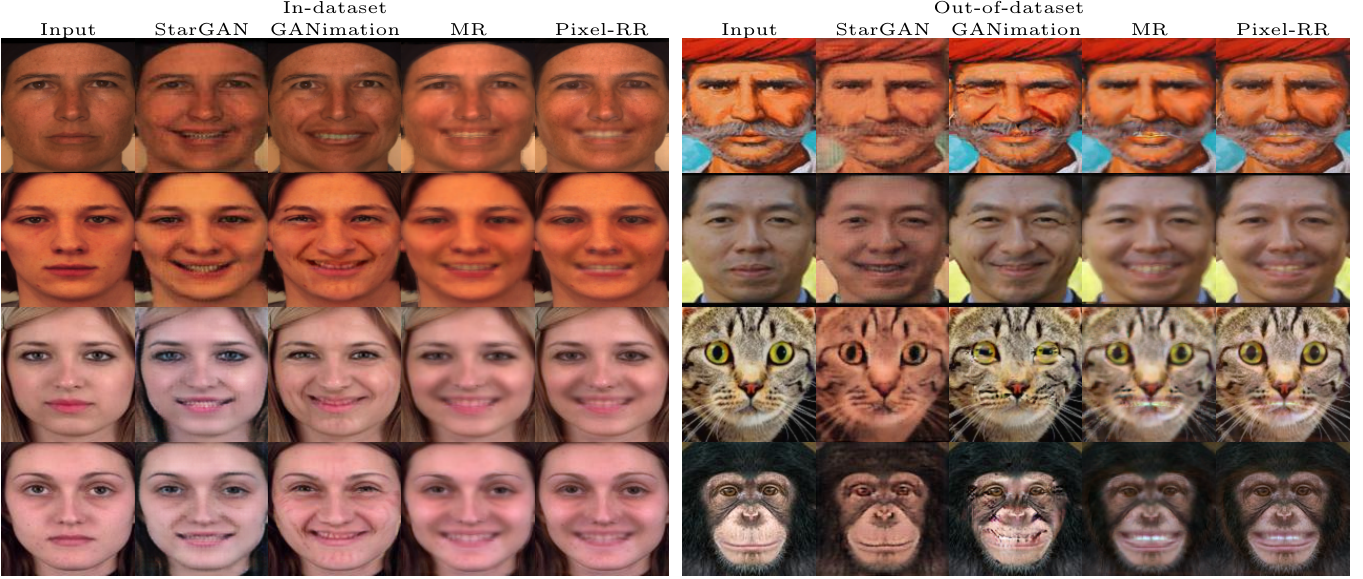}
    \caption{Comparison of Pixel-based ridge regression (Pixel-RR)  with state-of-the-art facial expression synthesis models, MR, StarGAN and GANimation. As shown in this figure, StarGAN is unable to induce plausible happy expression on in-dataset and out-of-dataset images (column 2 and 7, rows 1-4). GANimation successfully induces happy expression on in-dataset and out-of-dataset human faces. However, it also introduces noticeable artifacts around the eyes and mouth region (column 3, rows 2, 3 and 4; column 8, row 1). MR alleviates artifacts and generates plausible happy expression, but the synthesized images suffer from apparent blurring effects.
    Pixel-RR effectively transforms input expressions into the target expressions while preserving a person's identity and generates sharper images among all methods. Also, Pixel-RR preserves subtle details of the face. StarGAN and GANimation were not able to induce happy expression on animal faces while MR and our proposed method generalize well even on animal faces. Please zoom in for details.} 
    \label{fig:main_figure}
\end{figure*}

While capable of synthesizing photo-realistic facial expressions, existing GAN-based models are limited in two key aspects. First, these GAN-based methods generate photo-realistic results 
as long as the testing data distribution is similar to training data distribution or in-dataset images. Even the most successful expression synthesis methods such as StarGAN \cite{choi-2017} and GANimation \cite{pumarola2018ganimation} produce artifacts on out-of-dataset images of humans, paintings and animal faces as can be seen in Figure \ref{fig:main_figure}. 
It seems GANs have been inherently biased toward the training data distribution, since they are designed to replicate the training data distribution. 
Second, these GAN-based methods require thousands of images during training. The quality of results generated by GANs significantly degrades when trained on smaller datasets \cite{khan2019masked}. So, they do not support generalization with a few hundred images. 
Finally, GANs require higher computational and storage resources at testing time. Due to these constraints, it becomes challenging to deploy these models on resource-constrained devices and embedded systems which often have limited memory and processing power \cite{murshed2020machine}.

\begin{figure}[h]
    \centering
    \includegraphics[scale=0.4]{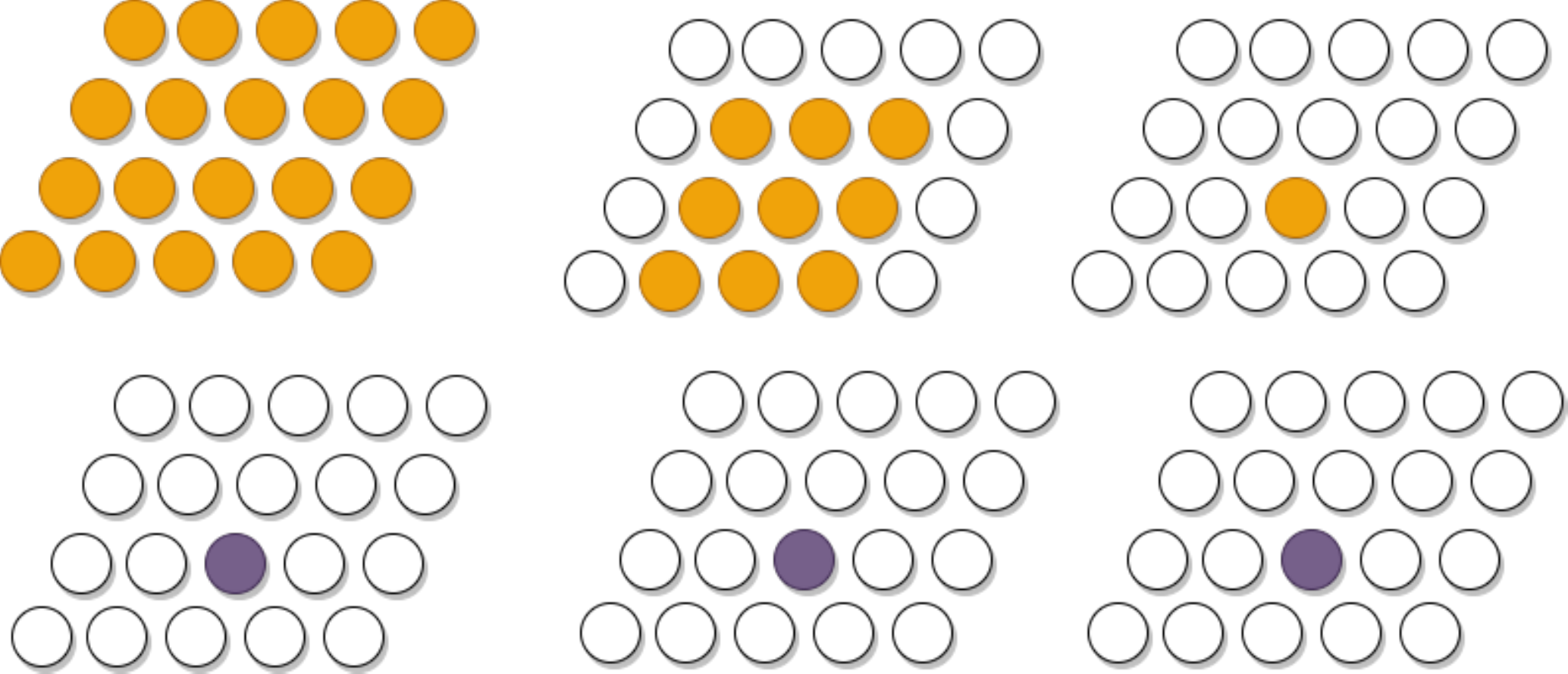}
    \caption{(Left to Right) Comparison between regression, masked regression and our proposed,  pixel regression. \textbf{Left:} One output pixel looks at all input pixels (Regression). \textbf{Middle:} Recently proposed MR \cite{khan2019masked} considers a local set of input pixels to make one output pixel (Masked Regression). The selection of pixels is fixed. 
    \textbf{Right:} Pixel Regression observes only one input pixel (Ours).}
    \label{fig:proposed_method}
\end{figure}

In an attempt to address these limitations,  we propose an alternative pixel-based approach to solve the facial expression synthesis problem. The key insight is derived from the work of Khan et al. \cite{khan2019masked} who showed that
facial expressions mostly constitute localized changes of the input image. Motivated by this fact, our proposed method considers only one input pixel to produce an output pixel. This is in contrast to traditional linear regression or neural network based models that look at all input image pixels. The difference between traditional methods and our proposed method is illustrated in Figure \ref{fig:proposed_method}. This pixel-based mapping significantly decreases the computational and space complexity of the proposed model by reducing the number of parameters.  
To capture complex, non-linear characteristics of facial expression mappings, we also show how kernel regression \cite{shi2019face} can be exploited.
At the cost of increased model size, this step can help to achieve sharper and more plausible synthesis of expressions. 
Linear regression or vanilla neural-network based methods contain millions of parameters due to being
fully connected.
To avoid overfitting in such a network large datasets are required for training. However, for FES large paired datasets are not readily available. 
One solution of this problem is to minimize the number of parameters by employing sparsity. Sparsity ensures that each output pixel observes only a few input pixels. Sparse models contain fewer parameters, generalize well and can be trained using smaller datasets. 

Facial expressions usually change localized regions of an image. This idea is exploited for masked regression (MR) \cite{khan2019masked}, that employs local receptive fields to observe a local patch of the input image in order to obtain one pixel of the synthesized image. We extend their local receptive field idea to achieve maximal sparsity and ultimate locality. We contend that there is no need to look at all pixels of a local input patch.  We call this approach \emph{pixel-based regression} which looks at only one fixed pixel of the input image to produce a pixel of the output image. 
The idea of looking at only one input-pixel enforces maximum sparsity as well as locality. 
Our method takes input and target image pairs and learns to transform the input expression into the target expression by employing pixel-based ridge regression (Pixel-RR). Then we also show that this idea can be extended to capture non-linear characteristics of facial images by using pixel-based kernel regression (Pixel-KR). We show that our proposed maximally localized and sparse method can successfully synthesize facial expressions and improve generalization performance. 
Furthermore, we formulate a convex objective function that achieves a global minimum and demonstrates superior performance as compared to deep generative models. Moreover, it incurs a very low computational cost and can be easily embedded in mobile devices. To the best of our knowledge, this is the lightest facial expression synthesis model that can be deployed in mobile devices and embedded systems. Table \ref{tab:parameters} shows that our proposed model (Pixel-RR) contains two orders of magnitude fewer parameters compared to state-of-the-art GANs.

\begin{table*}[ht]
    \centering
    \caption{Comparison of different FES models sizes. StarGAN and GANimation cannot be utilized in resource-constrained devices due to a large number of parameters. The proposed Pixel-RR model is two orders of magnitude smaller than deep GANs. It is also smaller than the recent linear regression based MR model while synthesizing sharper facial details (see Figure 4).}
    \begin{tabular}{cccccc} \hline
    Parameters  & StarGAN \cite{choi-2017} 
    &  GANimation \cite{pumarola2018ganimation}
    & MR \cite{khan2019masked}
    & Pixel-KR & Pixel-RR \\ \hline
    x$10^4$ & 850 & 850 & 16.2 & 655 & \textbf{3.28} \\
    \end{tabular}
    \label{tab:parameters}
\end{table*}

The main contributions of this work can be summarized as follows.
\begin{itemize}
    \item We have introduced the first pixel-based ridge regression method to solve the facial expression synthesis problem and have presented a kernel-based extension as well.
  
    \item Compared to state-of-the-art GAN-based models, the proposed method generalizes much better for a variety of out-of-dataset images.  
   
    \item 
    The proposed model is two orders of magnitude smaller than GAN-based models that makes it suitable for mobile devices and embedded systems.
  
\end{itemize}

\section{Related Work}
In the past few years, GANs have emerged as a powerful class of generative models for image-to-image translation problem. GANs exhibit an astonishing capability to generate photo-realistic and sharp images as opposed to blurry images generated via other traditional loss functions such as mean-squared error. GAN \cite{goodfellow-2014} consists of two neural networks that compete against each other to learn a minimax objective function. The first network, called generator $(G)$, tries to learn the data distribution while the other network, called discriminator $(D)$, tries to discriminate between real (coming from real data) and fake (coming from generator $G$) images. GAN  extension  to  generate  images  on  some  condition called conditional  Generative Adversarial  Network (cGAN) \cite{mirza2014conditional}. Due to the development of cGAN, image-to-image translation has achieved compelling results \cite{isola-2016, zhu-2017, choi-2017, karras2017progressive}. The Pix2pix \cite{isola-2016} framework uses cGAN and $\ell_1$ loss function to learn the mapping between paired input and output images. For unpaired image-to-image translation, cycleGAN and other frameworks \cite{kim-2017, liu-2017} have been proposed to learn the mapping in an unsupervised way. Conditional GANs have also been exploited to synthesize photo-realistic facial expressions. The most successful expression synthesis networks proposed in recent years are StarGAN \cite{choi-2017} and GANimation \cite{pumarola2018ganimation}. StarGAN, a multi-domain image-to-image translation network, has been proposed for facial expression synthesis among multiple domains using discrete labels. GANimation \cite{pumarola2018ganimation} is a facial action transfer method that employs GAN to synthesize facial expressions conditioned on action units. 
Although these GANs generate photo-realistic images, they also employ deep networks that entail large computational and storage overheads. Moreover, the performance of these GAN-based methods is not satisfactory when trained on small datasets. 
Recently, Khan et al. \cite{khan2019masked} proposed a linear regression-based method that addresses the above defined limitations of GANs. They observed that expressions mostly constitute local changes. With this motivation, they introduced masked regression to consider only local and sparse regions of images. However, the results of masked regression suffer from blurring effects. We extend their idea to learn maximally localized and maximally sparse receptive fields. This maximally localized and sparse receptive field idea reduces blurriness and enhances facial local details of the image. To improve the quality of results, we also employed kernel regression to capture non-linear characteristics of the face images.

\section{Pixel-based Regression}
The FES problem can be modeled as ridge regression (RR) whereby output faces are compared with target faces. Let $X \in \mathcal{R}^{N \times D}$ and $T \in \mathcal{R}^{N \times K}$ be the design and response matrices of $N$ input faces (vectorized in $\mathcal{R}^D$) and target faces (vectorized in $\mathcal{R}^K$) respectively. The objective function to learn this mapping via ridge regression can be written as 
\begin{equation}
    \centering
    E(W)= \frac{1}{2} \Vert WX^T-T^T \Vert_F^2 + \frac{\lambda}{2} \Vert W \Vert_F^2
    \label{eq:ridge_reg}
\end{equation}
where $W \in \mathcal{R}^{K \times D}$ is a transformation weight matrix and $\lambda>0$ is a regularization hyper-parameter that controls overfitting. The unique global minimizer for Eq. \ref{eq:ridge_reg}  can be computed analytically as
\begin{equation}
     W = \left[(X^TX + \lambda I)^{-1}X^T T \right]^T
     \label{eq:ridge_reg2}
\end{equation}

Facial expressions usually constitute local instead of global changes in the input image. For example, the transformation from neutral to happy mostly affects the regions around the eyes, nose and mouth to induce happy expression. It is therefore unnecessary to observe all input pixels while synthesizing a single output pixel. It should be sufficient to observe a relevant portion of the input image. We propose to observe only one input pixel for producing an output pixel. Let the input and target image pixels located at position $p$ in the $n$-th training pair be denoted as $x_p^n \in \mathcal{R}$ and $t_p^n \in \mathcal{R}$, respectively. Taking into account the pixel position, the entire training set can be divided into $K$ groups $\{\mathbf{x}_p, \mathbf{t}_p\}_{p=1}^K$, where $\mathbf{x}_p = [x_p^1, x_p^2, \dots, x_p^N] \in \mathcal{R}^{1 \times N}$ and $\mathbf{t}_p = [t_p^1, t_p^2, \dots, t_p^N] \in \mathcal{R}^{1 \times N}$ represent the corresponding input and target training pixel values at position $p$ for all $N$ training pairs. The objective function for pixel-based ridge regression (Pixel-RR) can be written as,

\begin{equation}
    \centering
    E^{}(w_p, b_p)= \frac{1}{2} \Vert w_p\mathbf{x}_p+b_p\mathbf{1} -\mathbf{t}_p \Vert_2^2 + \frac{\lambda}{2} (w_p^2 + b_p^2)
    \label{eq:pixel_ridge_reg}
\end{equation}

where scalars $w_p$ and $b_p$ are learnable weight and bias values that transform input pixels at position $p$ into output pixels and $\mathbf{1}$ is a vector of ones of size $1\times N$. . 
The partial derivatives of Eq. \ref{eq:pixel_ridge_reg} can be written as,
\begin{align}
    \frac{\partial E^\text{}(w_p, b_p)}{\partial w_p} &= \left(w_p \mathbf{x}_p + b_p\mathbf{1} - \mathbf{t}_p\right)\mathbf{x}_p^T + \lambda w_p = 0 \\
    \frac{\partial E^\text{}(w_p, b_p)}{\partial b_p} &= \left(w_p \mathbf{x}_p +b_p\mathbf{1} - \mathbf{t}_p\right)\mathbf{1}^T + \lambda b_p = 0 
\end{align}
from which the unique global minimizers for Eq. \ref{eq:pixel_ridge_reg} can be computed analytically as
\begin{align}
    \begin{bmatrix}
    w_p\\b_p
    \end{bmatrix}
    &=
    \begin{bmatrix}
    \mathbf{x}_p\mathbf{x}_p^T+\lambda & \mathbf{1}\mathbf{x}_p^T
    \\
    \mathbf{1}\mathbf{x}_p^T & N+\lambda
    \end{bmatrix}^{-1}
    \begin{bmatrix}
    \mathbf{t}_p\mathbf{x}_p^T
    \\\mathbf{t}_p\mathbf{1}^T
    \end{bmatrix}
    \label{eq:pixel_linear_regression}
\end{align}

Since linear regression cannot effectively capture complex relationships between inputs and outputs.
The proposed pixel-based mapping idea can be extended by using kernel regression. We can project input pixels $\mathbf{x}_p$ into a higher-dimensional Reduced Kernel Hilbert Space (RKHS) as 
\begin{align}
    \phi(\mathbf{x}_p) &= \begin{bmatrix}\phi(x_p^1)& \phi(x_p^2)& \dots & \phi(x_p^N)\end{bmatrix} \in \mathcal{R}^{d \times N}
\end{align}

The objective function pixel-based regression from this new RKHS to target values can be written as
\begin{equation}
    \centering
    E^{}(\mathbf{w}_p^\phi)= \frac{1}{2} \Vert \mathbf{w}_p^\phi \phi(\mathbf{x}_p)-\mathbf{t}_p \Vert_2^2 + \frac{\lambda}{2} \Vert \mathbf{w}_p^\phi\Vert^2
    \label{eq:kernel_ridge_reg}
\end{equation}
where $\mathbf{w}_p^\phi \in \mathcal{R}^{1 \times d}$ represents a learnable mapping from the RKHS to targets. 

According to the representer theorem, the kernel regression weights $\mathbf{w}_p^\phi$ can be represented by a finite linear combination of the input vector $\phi(\mathbf{x}_p)$ as
\begin{equation}
    \mathbf{w}_p^\phi = \mathbf{c}_p \phi(\mathbf{x}_p)^T 
\end{equation}
where $\mathbf{c}_p \in \mathcal{R}^{1 \times N}$ is the projection vector containing the coefficients of the linear combination. Thus, the optimization in Eq. \ref{eq:kernel_ridge_reg} can be reformulated to compute the projection matrix $\mathbf{c}_p$. The objective function for pixel-based kernel regression (Pixel-KR) can be written in terms of $\mathbf{c}_p$ as
\begin{align}
   E(\mathbf{c}_p) &= \frac{1}{2} \Vert \mathbf{c}_p \phi(\mathbf{x}_p)^T\phi(\mathbf{x}_p) - \mathbf{t}_p \Vert^2_2 + \frac{\lambda}{2} \Vert \mathbf{c}_p \phi(\mathbf{x}_p)^T \Vert^2_2 
   \\
   &=\frac{1}{2} \Vert \mathbf{c}_p K_p - \mathbf{t}_p \Vert^2_2 + \frac{\lambda}{2} \mathbf{c}_p K_p \mathbf{c}_p^T
   \label{eq:kernel_ridge_regresion1}
\end{align}
where $K_p=\phi(\mathbf{x}_p)^T\phi(\mathbf{x}_p) \in \mathcal{R}^{N \times N}$ is the kernel matrix that computes the non-linear similarity between $x_p^i$ and $x_p^j$ via the kernel function $k(x_p^i,x_p^j)=\phi(x_p^i)^T\phi(x_p^j)$
in the RKHS. We have used the Guassian kernel
\begin{align}
    k(x_p^i,x_p^j) &= \exp\left(\frac{-(x_p^i -x_p^j)^2}{2\sigma^2}\right) 
    \label{eq:kernel_function}
\end{align}

The gradient of Eq. \ref{eq:kernel_ridge_regresion1} w.r.t $\mathbf{c}_p$ can be written as:
\begin{align}
     \frac{\partial E^\text{}(\mathbf{c}_p)}{\partial \mathbf{c}_p} &= \left(\mathbf{c}_p K_p - \mathbf{t}_p\right)K_p + \lambda \mathbf{c}_p K_p = \mathbf{0}
     \label{eq:kernel_ridge_regresion3}
\end{align}
where we have used the fact that the kernel matrix $K_p$ is symmetric.  
The optimal projection matrix $\mathbf{c}_p$ can then be computed as
\begin{align}
     \mathbf{c}_p &= \mathbf{t}_p(K_p + \lambda I)^{-1} 
     \label{eq:kernel_ridge_regresion4}
\end{align}
Given an input pixel $x$ at testing time, the corresponding output pixel $y$ can be computed as
\begin{equation}
\begin{aligned}
    y &= w_p^\phi \phi(x) \\
        &= \mathbf{c}_p \phi(\mathbf{x}_p)^T \phi(x) \\
        &= \mathbf{t}_p(K_p + \lambda I)^{-1} \kappa(x)
\end{aligned}
\end{equation}
where the kernel vector $\kappa(x)$ is constructed as
\begin{align}
    \kappa(x) &= \begin{bmatrix}k(x_p^1,x) & k(x_p^2,x) & \dots & k(x_p^N,x)\end{bmatrix}^T
\end{align}

\section{Experiments and Results}
\begin{figure}[b]
    \centering
    \includegraphics[scale=1.8]{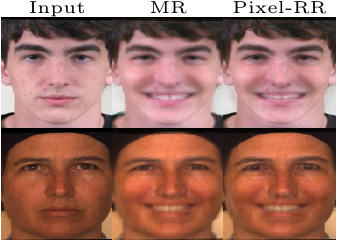}
    \caption{Visual comparison of MR and Pixel-RR. These results demonstrate that the Pixel-RR obtains sharper images with subtle facial details as opposed to MR. The results of MR suffer from smoothness artifacts. Pixel-RR suppresses the blurring effects, generates realistic images and retains local details of the face such as freckles, sparkle in the eyes and contours of the face image.}
    \label{fig:mr_vs_rr}
\end{figure}

\begin{figure*}[t]
    \centering
    \includegraphics[scale=1.8]{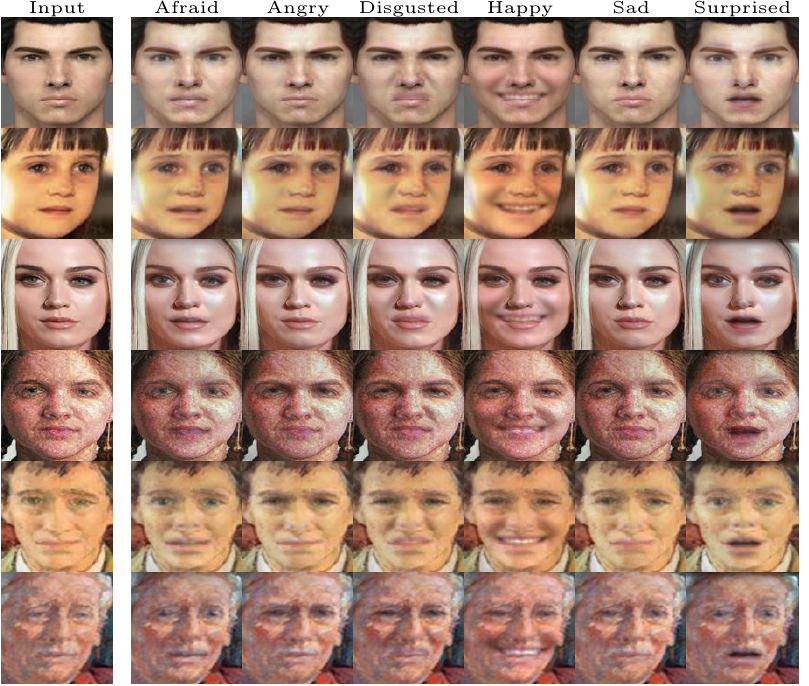}
    \caption{Qualitative results of Pixel-RR for facial expression synthesis on out-of-dataset images. 
    The proposed method successfully synthesized facial expressions on these out-of-dataset images. It can be seen that the proposed method performs well on portrait (last three rows) and non-frontal images (second and last row).}
    \label{fig:out_of_dataset}
\end{figure*}

The proposed method is evaluated on four well-known facial expression synthesis datasets (KDEF, Bosphorous, Radboud and CFEE). The Karolinska Directed Emotional Faces (KDEF) \cite{lundqvist-1998} dataset consists of 70 participants (35 Male, 35 Female) facial expressions face images. 
Bosphorous \cite{savran-2008} dataset consists of 105 participants facial expressions.  
Radboud faces dataset (RafD) \cite{langner-2010} consists of 67 participants facial expressions from five different angles and three different gaze directions (left, right and frontal). We used only frontal images with frontal gaze directions in our experiments. The Compound Facial Expressions of Emotion (CFEE) \cite{du-2014} database consists of 230 participants face expressions. 
To evaluate the generalization performance of our proposed method, some arbitrary celebrities and well-known people images, portraits and avatars are downloaded from the internet. These images, called out-of-dataset images, have significant variability in terms of background and pose.
We crop the face images into $128 \times 128$ size with faces in the center.
The regularization parameter $\lambda$ in Equation \ref{eq:pixel_linear_regression} and \ref{eq:kernel_ridge_regresion4} is set to $0.4$ after cross-validation. The parameter $\sigma$ in the Gaussian function in Eq. \ref{eq:kernel_function} is cross-validated and set to $3$ for neutral-to-happy mapping while for other expressions the sigma value is also cross-validated and set to $10$. The pre-trained GANimation model is used to generate GANimation results for comparison while StarGAN, MR, Pixel-RR and Pixel-KR are trained on the 400 images from all these datasets.

\subsection{Results on in-dataset and out-of-dataset images}
We compare our proposed method against state-of-the-art models including StarGAN \cite{choi-2017}, GANimation \cite{pumarola2018ganimation} and MR \cite{khan2019masked}. Qualitative results on in-dataset and out-of-dataset images can be seen in Figure \ref{fig:main_figure} and \ref{fig:out_of_dataset}.
It can be seen that StarGAN and GANimation generate quite convincing happy expression on these images. However, these methods also induce artifacts around the eyes and mouth regions. It must be noted that GANimation sometimes also fails to preserve a person’s identity as can be seen in the left panel of Figure \ref{fig:main_figure} (column 3, rows 1, 3 and 4). MR synthesizes good happy expressions as compared to  state-of-the-art GANs but the synthesized images also suffer from blurriness. Pixel-RR produces sharper results and induces plausible expressions on the input images. 
The right panel in Figure \ref{fig:main_figure} presents results on out-of-dataset images. 
Results demonstrate that GANs are not able to synthesize happy expressions on animal faces while regression-based methods MR and Pixel-RR successfully induce happy expressions on animal faces as well. Results for the proposed Pixel-RR method are sharper than results for MR. This can also be seen from Figure \ref{fig:mr_vs_rr} which also demonstrates that Pixel-RR retains finer facial details such as the sparkle in the eyes. Figure \ref{fig:out_of_dataset} demonstrates that Pixel-RR generalized well for out-of-dataset images such as portraits and faces with varying head pose.

\begin{figure*}[h]
    \centering
    \includegraphics[scale=1.65]{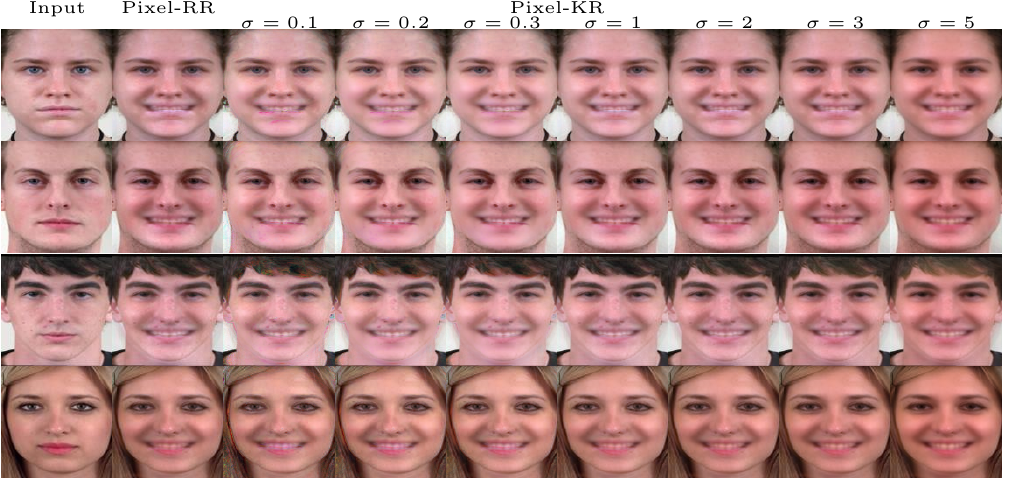}
    \caption{Results of Pixel-KR for varying $\sigma$. It can be observed that the results of Pixel-KR are highly sensitive to the optimal value of $\sigma$.}
    \label{fig:kernel_regression_with_different_sigma_values}
\end{figure*}

\subsection{Effects of $\sigma$ on the results of Pixel-KR}
As discussed and shown before, the idea of pixel-based mapping can be extended to capture the non-linear characteristics of facial images by using kernel-regression. However, Pixel-KR results are highly sensitive to choosing the optimal value of $\sigma$. To study the impact of $\sigma$ in the facial expression synthesis problem, we perform experiments by varying $\sigma$ from $0.1$ to $5$. Figure \ref{fig:kernel_regression_with_different_sigma_values} shows that values smaller than $1$ introduce artifacts on images whereas values greater than $1$ produce blurry images. It can be seen that Pixel-KR produces results similar to Pixel-RR when $\sigma = 1$. However, the number of parameters in Pixel-RR is much smaller as compared to Pixel-KR.

\subsection{Quantitative Evaluation Results}
In this subsection, we evaluate the quality of generated faces quantitatively.

\begin{table}[h]
    \centering
     \caption{User study to evaluate expressions and visual quality of generated images.}
    \begin{tabular}{c|c} \hline
      Model  & Neutral $\rightarrow$ Happy \\ \hline
      GANimation  &   0.26  \\
      MR   &          0.17  \\ 
      \textbf{Pixel-RR}  & \textbf{0.57}  \\ \hline
      \end{tabular}
    \label{tab:user_study}
\end{table}

\subsubsection{User Study}
A user study is conducted to evaluate the quality of the visual results of the proposed method. Ten input images are selected for evaluation. 
The human evaluators are asked to choose the best synthesized happy face image based on these three attributes: perceptual quality, expression realism and identity preservation. This study has been performed by 80 evaluators.
The results in Table \ref{tab:user_study} reinforce that our proposed regression-based method performs best among state-of-the-art facial expression synthesis methods. 

\subsubsection{Expression Classification Accuracy}
To quantitatively evaluate the out-of-dataset generalization performance of the proposed method, we use the pre-trained expression\footnote{\url{https://github.com/thoughtworksarts/EmoPy}} classifier to classify the synthesized images. A comparison of expression classification accuracy on the synthesized faces by our proposed method and other state-of-the-art facial expression synthesis methods is reported in Table \ref{tab:class_accuracy}. These results also support our claim that the proposed approach generalizes well even without deep networks and large training datasets. 

\begin{table}[ht]
    \centering
     \caption{Comparison of expression classification accuracy on faces generated by our proposed method and other state-of-the-art methods.}
    \begin{tabular}{cc} \hline
      Model  & Accuracy        \\ \hline
      GANimation  &  68\%      \\ 
      MR   &     84\%          \\ 
      \textbf{Pixel-RR} &  \textbf{85}\%  \\ \hline
      \end{tabular}
    \label{tab:class_accuracy}
\end{table}

\subsection{Performance of FES models on limited datasets}
GAN based FES models such as StarGAN and GANimation are trained on huge datasets. StarGAN employs CelebA dataset to synthesize facial images with different attributes while GANimation utilizes EmotionNet \cite{fabian2016emotionet} dataset for training. These datasets contain millions of images. However, such huge datasets are not readily available in all kinds of domains. 
Both pixel-based solutions perform well on in-dataset and out-of-dataset images while StarGAN is unable to produce realistic results (shown in Figure \ref{fig:main_figure}). It seems that StarGAN does not recover input image details and is not able to perform well with only few-hundred training images. This poor out-of-dataset generalization of GANs has also been discussed in \cite{khan2019masked}.
In contrast, our proposed method performs significantly 
well even with such a small dataset. It generates satisfactory results on in-dataset and out-of-dataset images.

\section{Conclusion}
We have presented a novel and simple pixel-based ridge regression model for facial expression synthesis that considers only one input pixel to produce an output pixel. Qualitative and quantitative results confirm that Pixel-RR produces promising results on in-dataset and out-of-dataset images as compared to state-of-the-art GAN models while requiring two orders of magnitude fewer parameters. Due to the smaller number of parameters, Pixel-RR can be easily deployed in mobile devices and embedded systems to synthesize realistic facial expressions. We have also shown that the pixel-based idea can be extended to capture non-linear characteristics of facial expression mappings by using kernel-regression. Results demonstrate that the proposed Pixel-RR and Pixel-KR methods effectively transform an input expression into a target expression while preserving identities and facial details. However, using the kernel-based method comes at the cost of increased number of parameters.  

\bibliographystyle{IEEEtran}
\bibliography{ref.bib}
\end{document}